\documentclass[11pt]{article}
\usepackage[utf8]{inputenc}
\usepackage{longtable}
\usepackage{blindtext,alltt}
\usepackage{listings}
\usepackage{xcolor}
\usepackage{graphicx}
\usepackage{hyperref}
\usepackage[margin=1.5in]{geometry}
\title{Patch and Shuffle: A Preprocessing Technique for Texture Classification in Autonomous Cementitious Fabrication}

\author{Jeremiah Giordani\\
Princeton University\\
\texttt{jg0037@princeton.edu}
}

\begin{document}

\maketitle

\begin{abstract}
Autonomous fabrication offers enormous potential to the field of civil engineering. However, such processes have significant challenges and limitations. One such drawback is that errors frequently arise during fabrication. When such abnormalities arise, most fabrication systems do not have a method to stop or adjust the print. However, in recent years, some fabrication systems have integrated computer vision systems to detect printing errors and adjust print settings when errors arise. This paper presents a novel computer vision architecture designed to improve texture classification in autonomous cementitious fabrication. Traditional approaches to texture classification in autonomous fabrication rely on full image analysis, which can cause classifiers to base classifications on the semantic or high-level content of images. However, these approaches may be less effective for detecting texture-based characteristics. To address this shortcoming, the proposed system uses a novel approach we call "patch and shuffle" Each input image is divided into smaller patches that are shuffled, before being passed into a classifier. Isolating smaller regions of the print images removes the overall semantic content in the image. This allows the classifier to focus on texture-based features instead of high-level content. In this paper, I develop the classification architecture and train it on a dataset containing images of cementitious fabrication. I then evaluate the performance of the classifier and show that it improves accuracy, compared to existing architectures. This work builds on the existing research surrounding computer vision texture classification systems and presents a novel architecture that future work could modify or extend. \footnote{This work was originally completed as a final project for CEE 374 at Princeton University.} 

\end{abstract}

\section{Background}  

Computer vision is a field of artificial intelligence focused on image or video analysis, extracting information from visual data \cite{computer_vision}. It employs a variety of algorithms to analyze visual inputs and perform a wide range of tasks, including prediction, classification, object recognition, and much more. 

Computer vision plays a critical role in autonomous fabrication, addressing challenges such as error detection, texture classification, and automated adjustment, \cite{review}, \cite{error1}, \cite{error2}, \cite{2020_paper}, \cite{adjustment}. Fabrication systems that integrate computer vision allow for precise monitoring and control of the manufacturing process, enabling the identification of defects and print quality. Such systems can create an automated printing pipeline that allows for real-time adjustments, improving fabrication reliability \cite{system1}. 

Applications of computer vision in autonomous fabrication throughout research can be grouped into several categories. Some of the most common use-cases are as follows. 

\begin{enumerate}  
    \item \textbf{Error Detection:} Identifies deviations from expected printing patterns during the fabrication process \cite{error1}, \cite{error2}, \cite{error3}, \cite{error4}, \cite{error5}.
    \item \textbf{Real-time Printing Correction:} Predicts print adjustments and automates parameter correction \cite{correction}.
    \item  \textbf{Texture Classification:} Classifies extruded material into various texture categories \cite{2020_paper}.
    \item \textbf{Quality Monitoring:} Analyzes the print to verify consistency with fabrication specifications \cite{quality}.
\end{enumerate}  

This paper focuses on texture classification, with particular application to cementitious fabrication.

\subsection{Texture Classification}

Texture classification refers to the process of using computer vision during an additive manufacturing process in order to classify and identify material texture \cite{2020_paper}. The technique uses machine learning to identify and differentiate between various categories of extruded material. For cementitious fabrication, this process focuses on classifying images of extruded material based on its fluid content.

Texture classification processes are critical because the texture of extruded material directly impacts the structural integrity, mechanical properties, and overall print quality \cite{cement-sensitive}. Inconsistent textures can indicate improper extrusion rates, nozzle blockages, material inconsistencies, or simply material variations that arise naturally during fabrication. 

Detecting these issues early allows for corrective actions in real-time, which reduces waste, improves the structural consistency of the final product, and potentially saves a fabrication from failure \cite{cement-error-detection}. 

\subsection{Cementitious Manufacturing}

Cementitious manufacturing is the processes of using cement-based materials in additive manufacturing. This process is widely applicable to a variety of cases in construction and civil engineering, ranging from smaller components to entire architectures \cite{cement-general}. Current research is aimed at increasing its scalability and efficacy, with the goal of unlocking its enormous future potential in large-scale construction \cite{cement-scale}.

However, cementitious manufacturing has significant drawbacks and limitations. Such fabrication processes are highly susceptible to errors during the printing process. The cement rheological properties, such as plasticity, viscosity, elasticity play an important role in the overall quality and structural integrity of the print \cite{cement-sensitive}. When these properties deviate from ideal set points, it can result in defects and structural weaknesses.  

Automated systems that can detect the rheological properties of the cement paste during extrusion offer enormous potential in designing systems that can mitigate these issues. Texture classification, in particular, is a potential component to systems aimed at addressing these challenge \cite{2020_paper}. Images of the print can be used to analyze the surface characteristics and classify the rheological state of the material. Such classification can enable manual, and, importantly, automated adjustments to printing parameters in real-time.

\section{Framework and Approach}

In this paper, we focus on a particular application of computer vision in cementitious manufacturing. It builds on existing work that aims to classify the texture of extruded cement, classifying print images in one of four categories, "fluid", "good", "dry", or "tearing".

For this problem, we design a novel computer vision classification architecture. The goal is to test our hypothesis that texture-based classification will work better by pre-processing the images through a transformation we call "patch and shuffle."

\begin{figure}[h]
    \centering
    \includegraphics[width=0.5\textwidth]{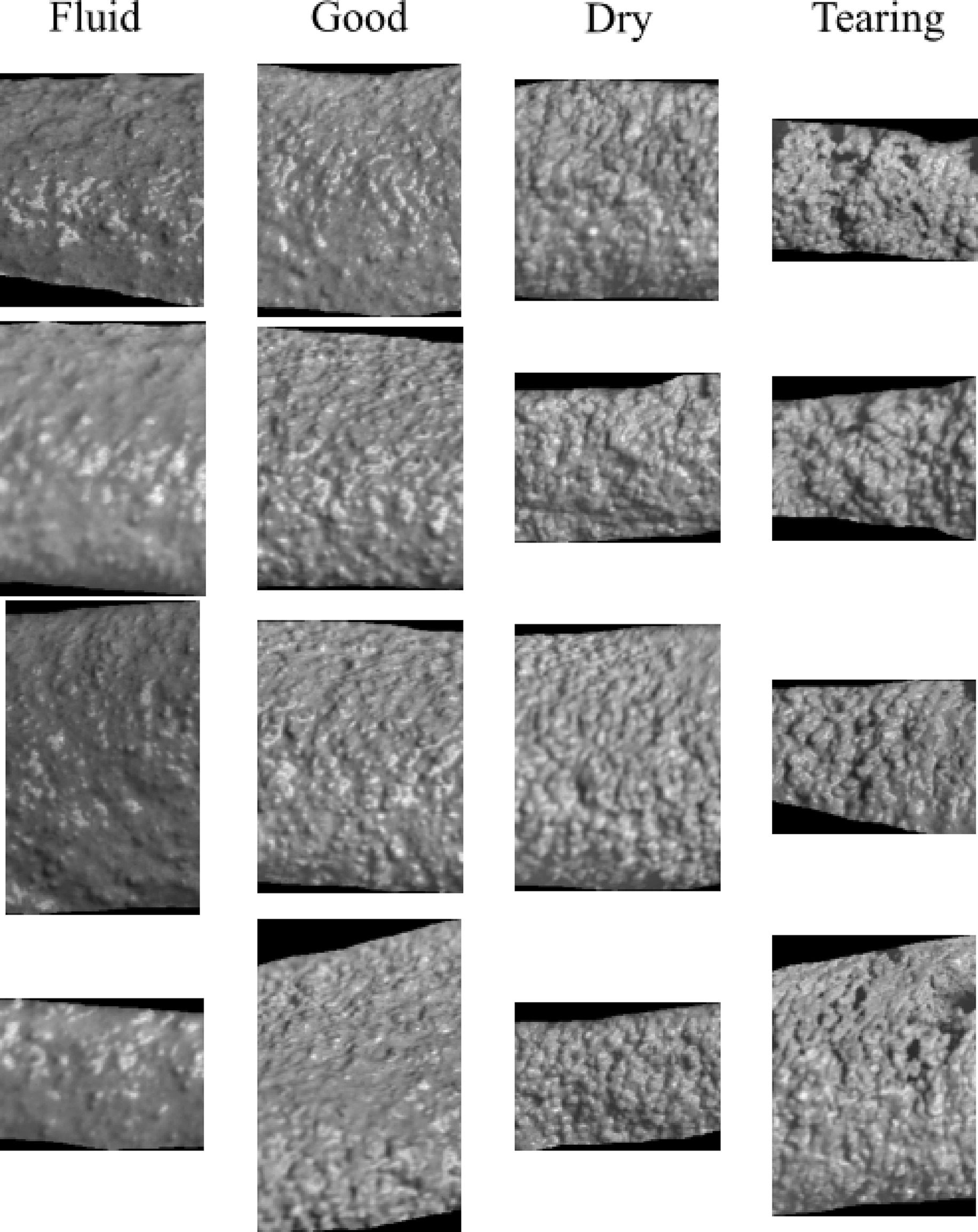}
    \caption{Texture Windows - Unmodified from \cite{2020_paper}}
    \label{fig:texture-windows}
\end{figure}

\subsection{Problem Definition}

In the 2020 paper "Inline monitoring of 3D concrete printing using computer vision" \cite{2020_paper}, the authors develop a framework for a computer vision system with wide ranging functionality and use cases for cementitious printing. One of these steps involves texture classification, which is the focus of this paper. 

A "texture window" is a small, cropped image of extruded cement. Each texture window is classified into four different categories, "fluid", "good", "dry", and "tearing". Figure \ref{fig:texture-windows} shows a set of example texture windows, grouped by their classification. 

\begin{figure}[h]
    \centering
    \includegraphics[width=0.8\textwidth]{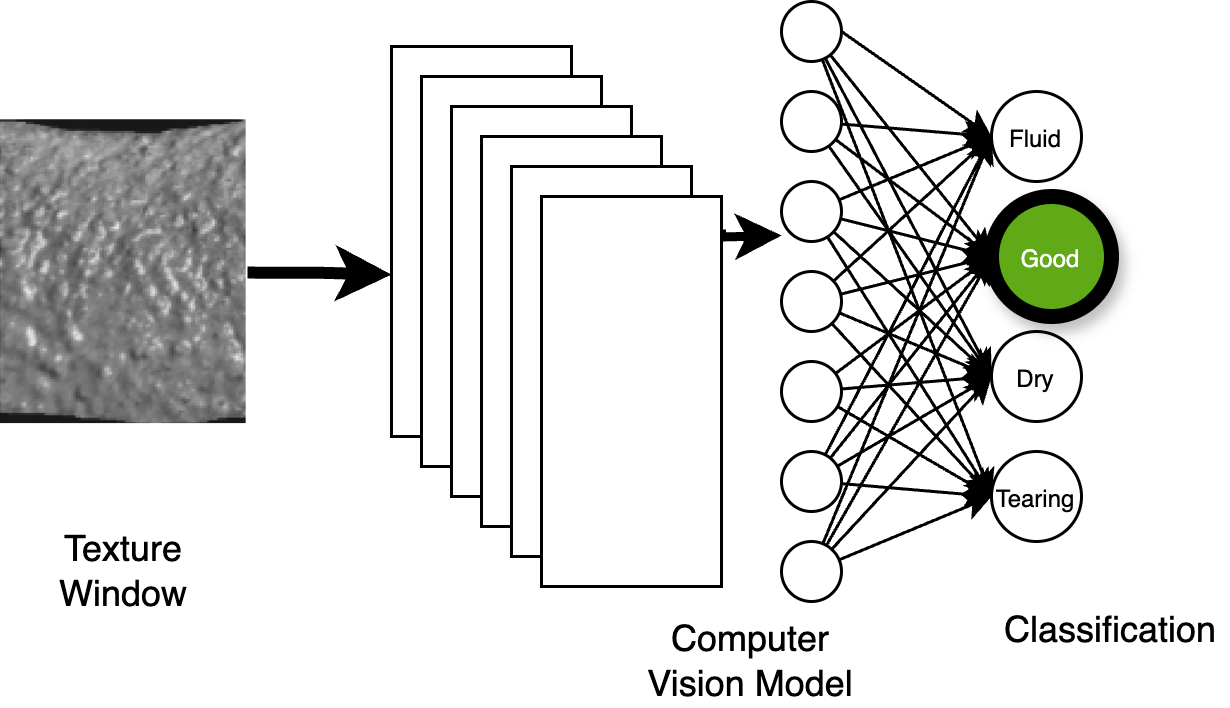}
    \caption{Classification Pipeline}
    \label{fig:class-pipeline}
\end{figure}

Computer vision is used in this system in order to process texture windows as input and generate classifications as output. Figure \ref{fig:class-pipeline} offers an illustration of the standard classification pipeline. Texture windows are taken as input and classifications are outputted.

\subsection{Proposed Approach}

\begin{figure}[h]
    \centering
    \includegraphics[width=0.5\textwidth]{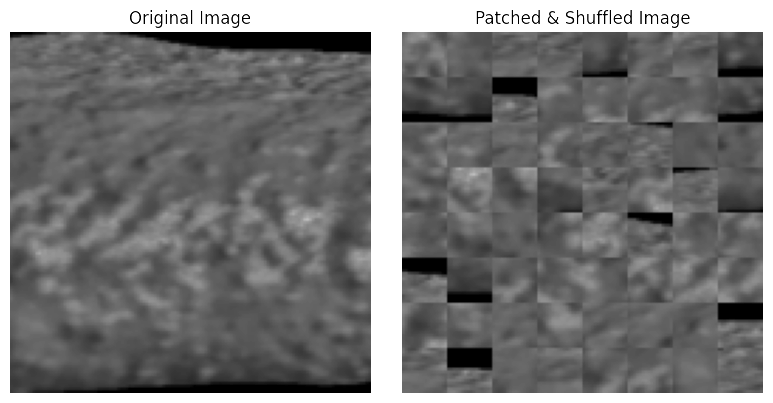}
    \caption{Original image (left) compared to transformed image (right) }
    \label{fig:patch-and-shuffle}
\end{figure}

Our approach is to transform the data prior to passing it into the model, while holding the model itself and its architecture constant. Figure \ref{fig:patch-and-shuffle} shows how data is transformed in our approach. The original image is placed next to the transformed image that underwent "patch and shuffle", the transformation that our novel approach introduces. 

\begin{figure}[h]
    \centering
    \includegraphics[width=0.8\textwidth]{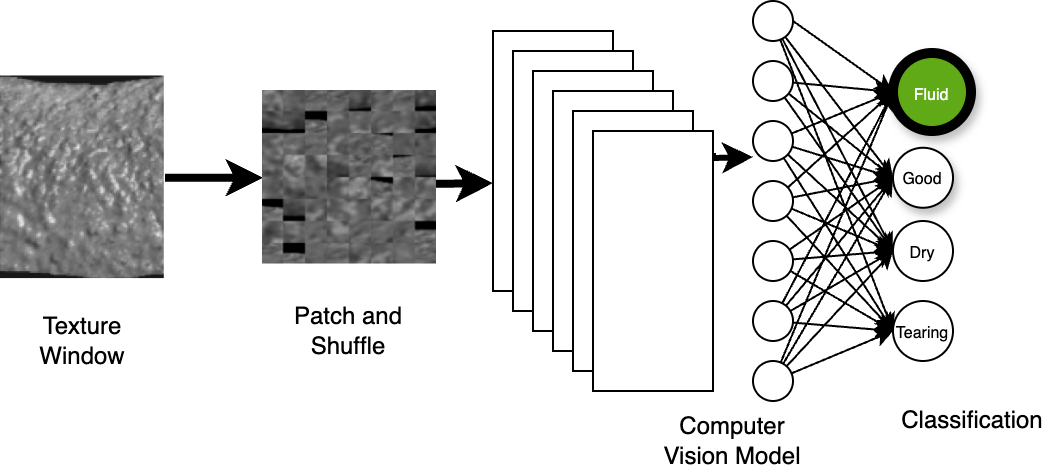}
    \caption{Proposed Classification Pipeline}
    \label{fig:our-class-pipeline}
\end{figure}

Our hypothesis is relatively straightforward: if input data is transformed by "patch and shuffle" prior to inputting it into the classifier, the classifications will be more accurate. Figure \ref{fig:our-class-pipeline} shows how our approach differs from the standard approach. 

\section{Experimental Design}

We designed the following experiment to test our hypothesis. The experimental design was carefully controlled to ensure that the results could be used to evaluate the efficacy of the patch and shuffle transformation.

\subsection{Dataset}

We used a dataset of labeled texture windows from cementitious prints from the 2020 paper referenced above \cite{2020_paper}. This \href{https://github.com/Sutadasuto/I3DCP}{dataset} is publically available on GitHub. Each texture window contains a label in a csv that maps images to labels. This dataset was generated and cleaned by the authors of the paper. 

\subsection{Classifier}

We developed an original \href{https://github.com/JeremiahGiordani/texture_classifier}{GitHub} repository that defines the model and training algorithms. The repository is publically available. 

\begin{itemize}
\item \textbf{Dataset Module:} This module handles dataset pre-processing and manages importing data into PyTorch dataloaders.  
\item \textbf{Model Architecture:} We use a pretrained ResNet-18 as the base classifier. The final fully connected layer was replaced with a linear layer producing four output classes, with a dropout probability of 0.5 for regularization. 
\item \textbf{Augmentations:} We apply the same data augmentations as used in the 2020 paper, applying random rotations, zooms, and changes in illumination.
\item \textbf{Patch and Shuffle} We implemented the patch and shuffle transformation, which divides the input image into smaller patches, randomly shuffles them, and reconstructs the shuffled image.  
\item \textbf{Training Algorithm:} We designed the training algorithm to allow configurable hyperparameters through script flags. The dataset was randomly split into training and testing sets. We used Cross-entropy loss as the optimization criterion, and an Adam optimizer. We implemented a testing loop to iterate over multiple epochs.

Configurable parameters are as follows:
\begin{itemize}
    \item Number of Epochs
    \item Learning Rate
    \item Weight Decay
    \item Patch Size
    \item Training/Testing Split
\end{itemize}
\end{itemize}

\subsection{Computing Clusters}

In order to train and test our classifiers, we used \href{https://researchcomputing.princeton.edu/systems/adroit}{Princeton's Adroit computing cluster}. Job scheduling was configured using Slurm. This setup enabled efficient training and testing, as well as an efficient method for hyperparameter search. By leveraging command line flags, we configured different jobs to run with different hyperparameters.

\subsection{Control and Experimental Groups}

To test our hypothesis, we trained, tested, and compared the results for two models

\begin{itemize}
\item \textbf{Control Model:} A standard classification pipeline, as represented in Figure \ref{fig:class-pipeline}. This trained the model on unmodified texture windows.
\item \textbf{Experimental Model:} A classification pipeline that transformed images through patch and shuffle prior to training the classifier. The model is represented in \ref{fig:our-class-pipeline} 
\end{itemize}

The setup outlined above enabled a direct comparison of classification accuracy between the standard pipeline and our proposed approach. When holding our model constant and only changing the image preprocessing, we could analyze the direct impact of patch and shuffle on the accuracy.

\section{Results}

We now compare the results of our approach with the results of the standard classifier. From hyperparameter search, we determined the optimal configuration to iterate of 30 epochs, using a learning rate of $5 \times 10^{-5}$, and a weight decay of $0.001$. The dataset was split into training and testing sets, where training used an Adam optimizer to minimize cross-entropy loss. Hyperparameters were fine-tuned using Slurm job scheduling on Princeton's computing clusters. Job outputs are linked in the Appendix.

The results for the best iteration and overall performance are listed in Table \ref{tab:results}

\begin{table}[h]
    \centering
    \caption{Comparison of Patch-and-Shuffle and Standard Classifiers}
    \begin{tabular}{|l|c|c|}
        \hline
        \multicolumn{3}{|c|}{\textbf{Best Iteration}} \\ \hline
        \textbf{Metric} & \textbf{Patch-and-Shuffle} & \textbf{Standard} \\ \hline
        Overall Accuracy (\%) & \textbf{90.64} & 72.46 \\ \hline
        "Fluid" Class Accuracy (\%) & 87.50 & 96.08 \\ \hline
        "Good" Class Accuracy (\%) & 98.82 & 61.34 \\ \hline
        "Dry" Class Accuracy (\%) & 66.67 & 72.06 \\ \hline
        "Tearing" Class Accuracy (\%) & 100.00 & 75.00 \\ \hline
        \multicolumn{3}{|c|}{\textbf{Overall Results}} \\ \hline
        Average Accuracy (\%) & \textbf{86.84} & 68.71 \\ \hline
    \end{tabular}
    \label{tab:results}
\end{table}

\subsection{Observations}  

The findings support our hypothesis that preprocessing the data through patch and shuffle enhances the accuracy of texture classification. For the best iteration, the patch and shuffle approach outperformed the standard approach at \textbf{90.64\%} compared to \textbf{72.46\%}. 

Performance improvements were observed across multiple texture classes. The patch and shuffle approach achieved $98.82\%$ accuracy for "good" texture windows, and $100.00\%$ accuracy for "tearing" texture windows, both of which outperformed the standard classifier. However, "dry" texture windows showed a lower accuracy of $66.67\%$, indicating that patch and shuffle showed some sensitivity to particular texture variations. 

When averaged across all iterations, the patch and shuffle classifier achieved an overall accuracy of \textbf{86.84\%}, compared to an overall accuracy of \textbf{68.71\%} for the standard classifier. This shows a significant performance increase using the patch and shuffle approach, supporting our initial hypothesis. 

In addition, the patch and shuffle approach showed better generalization, with a lower disparity between training and testing accuracies compared to the standard approach. This supports the conclusion that the patch and shuffle approach reduces overfitting.

\subsection{Limitations}  

The resulted outlined above offer important insights, however, they should be taken in the context of certain limitations. 

\begin{enumerate}
    \item \textbf{Dataset:} The provided dataset, while thorough and consistent, is relatively small. Data augmentations are used to increase the dataset by a factor of 16. However, having a limited dataset should be noted in contextualizing the results.
    \item \textbf{Class Imbalance:} While classes are distributed roughly equally, there is some minor imbalance in the dataset. 
\end{enumerate}

These limitations are important in understanding and contextualizing the results. However, we still feel that the results support the conclusion regarding the efficacy of the patch and shuffle approach. Issues with the dataset impact the trained model, however, our experiment design was focused on comparing accuracies between models, not evaluating the performance of a single model. 

Despite these limitations, we argue that the controlled experimental design ensures that the observed difference are due to the patch and shuffle transformation. The results indicate that \textit{under identical conditions}, the patch and shuffle approach consistently outperformed the standard pipeline. The consistent performance gains suggest that this preprocessing technique is a valuable performance enhancement, even in suboptimal conditions. 

\section{Analysis}

In this section, we propose two potential explanations for the observed results. Here, we formulate theories regarding the underlying cause behind performance increases exhibited by the patch and shuffle approach.

\subsection{Factor 1: Reduced Reliance on Semantic Content}

The underlying architecture behind most computer vision models, including the one used in the standard classifier in this study, is based on convolutional neural networks (CNNs). These networks are highly adaptable frameworks, capable of identifying patterns in images and extracting semantic content contained in visual data. 

Semantic content refers to the high-level information in an image. In other words, it refers to the objects or relationship between objects in an image. For example, in an image with a tree, the semantic content refers to the presence of the tree itself as an identifiable object in the image.

CNNs excel at capturing this type of information. This makes them highly effective and accurate in tasks like object detection and image classification, where the semantic content directly correlates with and defines the desired output. 

However, in texture classification, the task fundamentally differs from tasks reliant on semantic content. In texture classification, the texture is defined in the low-level details of an image. Here, the arrangement and intensity of surface patterns define the desired output. High-level semantic features can be "distracting" to the classifier, potentially leading to incorrect classifications if the model relies to heavily on these high-level features. 

The patch and shuffle approach addresses this challenge by destroying the semantic content of the texture windows. When it transforms the image by dividing it into smaller patches and shuffling them randomly, the overall structure and context of the image is disrupted. This forces the classifier to focus exclusively on the micro-level details, which might include local patterns and low-level texture gradients, when making classifications. The shift in focus from high-level to low-level features in making its decision is a potential factor in its performance enhancement. 

\subsection{Factor 2: More Robust Augmentation}

The increase in accuracy with the patch and shuffle approach cal also be interpreted as a result of more robust data augmentations. In machine learning, data augmentation techniques are used to improve model generalization by introducing variations in the training data. Common augmentations include rotations, translations, or crops. 

The patch and shuffle approach can be thought of as a more aggressive and disruptive data augmentations. Each shuffled texture window represents a new variation of the original image, while still preserving the low-level texture features. The increased variability of data prevents the classifier from overfitting to specific training data, increasing its generalizability and improving its performance on unseen test data. 

Additionally, by forcing the model to learn a wider range of patterns, the patch and shuffle transformation behaves as a regularization mechanism. This can also mitigate overfitting issues and improve generalization. 

\section{Future Directions}  

This section will explore potential applications of these findings, and outlines promising avenues for future research.

\subsection{Extending this Work}   

Our findings demonstrate the potential performance improvements of the patch and shuffle approach to improve accuracy in texture classification. This highlights several avenues for further exploration.

As outlined in the Limitations section, one drawback of this study is the dataset. While the dataset was effective for our experiments, future research could build on this work by evaluating our approach on larger and more diverse datasets. Future work could confirm or deny the robustness of this approach across different data distributions. 

It would also be valuable to experiment with this approach on other fabrication methods, such as autonomous plastic manufacturing. Such studies would be valuable in confirming or denying the extent to which this approach could generalize to materials with different surface characteristics. 

Another potential extension of this work could be in error detection. Experiments that test the patch and shuffle approach to classify errors during the printing process could lead to advancements in real-time monitoring systems in autonomous fabrication.

\subsection{Implications of this Work}  

The success of the patch and shuffle approach suggests that \textit{any} computer vision system designed to classify images based on low-level details may benefit from similar processing techniques. While our results indicate the efficacy of patch and shuffle in a limited context, the underlying cause of the performance enhancements imply that such a method might be useful in a broad range of tasks. 

In image classification tasks where high-level contextual information is less critical than low-level patterns, the findings from this paper indicate that destroying the semantic content in the images can improve the model performance. This insight can be widely integrated to various classification tasks. 

\subsection{Integrating the Findings}  

The findings from this paper could be integrated into cementitious fabrication systems. Such a system might integrate a computer vision model in order to perform texture classification in a direct feedback loop with adding accelerant into the material in the nozzle. 

If the extruded material becomes too fluid, the computer vision system could identify this and reduce the accelerant flow at the nozzle. Likewise, if it becomes too dry, it could increase the flow of accelerant. Such a system could improve the reliability and adaptability of cementitious fabrication, autonomously

\section{Conclusion}
The study outlined in this paper demonstrates the efficacy of a preprocessing technique we call "patch and shuffle" in enhancing texture classification accuracy for cementitious fabrication. The approach, which disrupts the semantic content in the texture windows, allows for classifiers to focus on low-level features. The results validate our hypothesis that transforming the data through patch and shuffle can enhance the performance of texture classifiers. 

The findings may extend beyond cementitious manufacturing, and imply broad applicability. Future research can build on this work to evaluate this approach on other types of fabrication and even entirely different domains. Integrating these findings into a cementitious fabrication system could improve the efficacy and scalability of additive manufacturing.

\newpage
\section*{Acknowledgements}
I would like to thank Professor Reza Moini for his guidance and support throughout the development of this work, which was originally completed as part of CEE 374 at Princeton University.
\nocite{*}

\lstset{ 
    basicstyle=\ttfamily\small,
    backgroundcolor=\color{lightgray!20},
    frame=single, breaklines=true,
    captionpos=b,
    showstringspaces=false
}

\newpage
\appendix
Below, we link training output files, as well as list the relevant outputs described above.

\subsection{Standard Approach Training Results}

The full \href{https://github.com/JeremiahGiordani/texture_classifier/blob/main/training_outputs/standard_train-2286793.out}{output file} can be found on the GitHub Repo. Truncated results are printed below.
\begin{lstlisting}[caption={Standard Training Output}, label={lst:classifier_output}]
Using STANDARD approach
Epochs: 30
Learning rate: 5e-05
Weight decay: 0.001
Using device: cuda
Using Cross Entropy Loss

[TRUNCATED]

Epoch [15/30]
                                                        
Train Loss: 0.0049 | Train Acc: 100.00%
  Train Class 0 Acc: 100.00%
  Train Class 1 Acc: 100.00%
  Train Class 2 Acc: 100.00%
  Train Class 3 Acc: 100.00%
Test Loss: 0.8376 | Test Acc: 72.46%
  Test Class 0 Acc: 96.08%
  Test Class 1 Acc: 61.34%
  Test Class 2 Acc: 72.06%
  Test Class 3 Acc: 75.00%
Average Accuracy: 68.37%
Model saved.

[TRUNCATED]

Training complete.
Best Test Acc: 72.46%
Average Accuracy: 68.71%
\end{lstlisting}

\newpage
\subsection{Patch and Shuffle Approach Training Results}

The full \href{https://github.com/JeremiahGiordani/texture_classifier/blob/main/training_outputs/standard_patch_train-2286792.out}{output file} can be found on the GitHub Repo. Truncated results are printed below.
\begin{lstlisting}[caption={Patch and Shuffle Training Output}, label={lst:classifier_output}]
Using PATCH approach
Epochs: 30
Learning rate: 5e-05
Weight decay: 0.001
Using device: cuda
Using Cross Entropy Loss

[TRUNCATED]

Epoch [21/30]
                                                        
Train Loss: 0.1052 | Train Acc: 95.97%
  Train Class 0 Acc: 97.06%
  Train Class 1 Acc: 94.12%
  Train Class 2 Acc: 93.84%
  Train Class 3 Acc: 98.24%
Test Loss: 0.3477 | Test Acc: 90.64%
  Test Class 0 Acc: 87.50%
  Test Class 1 Acc: 98.82%
  Test Class 2 Acc: 66.67%
  Test Class 3 Acc: 100.00%
Average Accuracy: 86.57%
Model saved.

[TRUNCATED]

Training complete.
Best Test Acc: 90.64%
Average Accuracy: 86.84%
\end{lstlisting}

\end{document}